\documentclass[10pt,conference,twocolumn]{IEEEtran}
\usepackage{cite}
\usepackage{amsmath,amssymb,amsfonts}
\usepackage{algorithmic}
\usepackage{graphicx}
\usepackage{textcomp}
\usepackage{xcolor}
\usepackage[hyphens]{url}
\usepackage{fancyhdr}
\usepackage{hyperref}
\usepackage{xspace}
\usepackage{enumitem}
\usepackage{wrapfig}
\usepackage{xcolor}

\newcommand\camerareadyaddition{\color{black}}
\usepackage{caption}
\usepackage{subcaption}

\newcommand{\hpcayear}{2024}

\title{Beyond Efficiency: Scaling AI Sustainably}

\def\hpcacameraready{} 

\newcommand\hpcaauthors{Carole-Jean Wu, Bilge Acun, Ramya Raghavendra, Kim Hazelwood \\
\\
FAIR at Meta
\vspace{-0.25cm}
}

\author{
  \ifdefined\hpcacameraready
    \IEEEauthorblockN{\hpcaauthors{}}
      \IEEEauthorblockA{
      }
  \else
    \IEEEauthorblockN{\normalsize{HPCA \hpcayear{} Submission
      \textbf{\#\hpcasubmissionnumber{}}} \\
      \IEEEauthorblockA{
        Confidential Draft \\
        Do NOT Distribute!!
      }
    }
  \fi 
}

\begin{document}
\maketitle

\pagestyle{fancy}

\newcommand{\hpcaheight}{0mm}
\ifdefined\eaopen
\renewcommand{\hpcaheight}{12mm}
\fi

\begin{abstract}
Barroso’s seminal contributions in energy-proportional warehouse-scale computing launched an era where modern datacenters have become more energy efficient and cost effective than ever before. 
At the same time, modern AI applications have driven ever-increasing demands in computing, highlighting the importance of optimizing efficiency across the entire deep learning model development cycle. 
This paper characterizes the carbon impact of AI, including both operational carbon emissions from \textit{training} and \textit{inference} as well as embodied carbon emissions from datacenter construction and hardware manufacturing. We highlight key efficiency optimization opportunities for cutting-edge AI technologies, from deep learning recommendation models to multi-modal generative AI tasks. To scale AI sustainably, we must also go beyond efficiency and optimize across the life cycle of computing infrastructures, from hardware manufacturing to datacenter operations and end-of-life processing for the hardware.    
\end{abstract}

\section{From Warehouse Scale Computing to AI}

\begin{figure}
\vspace{0.25cm}
  \centering
  \includegraphics[width=0.9\columnwidth]{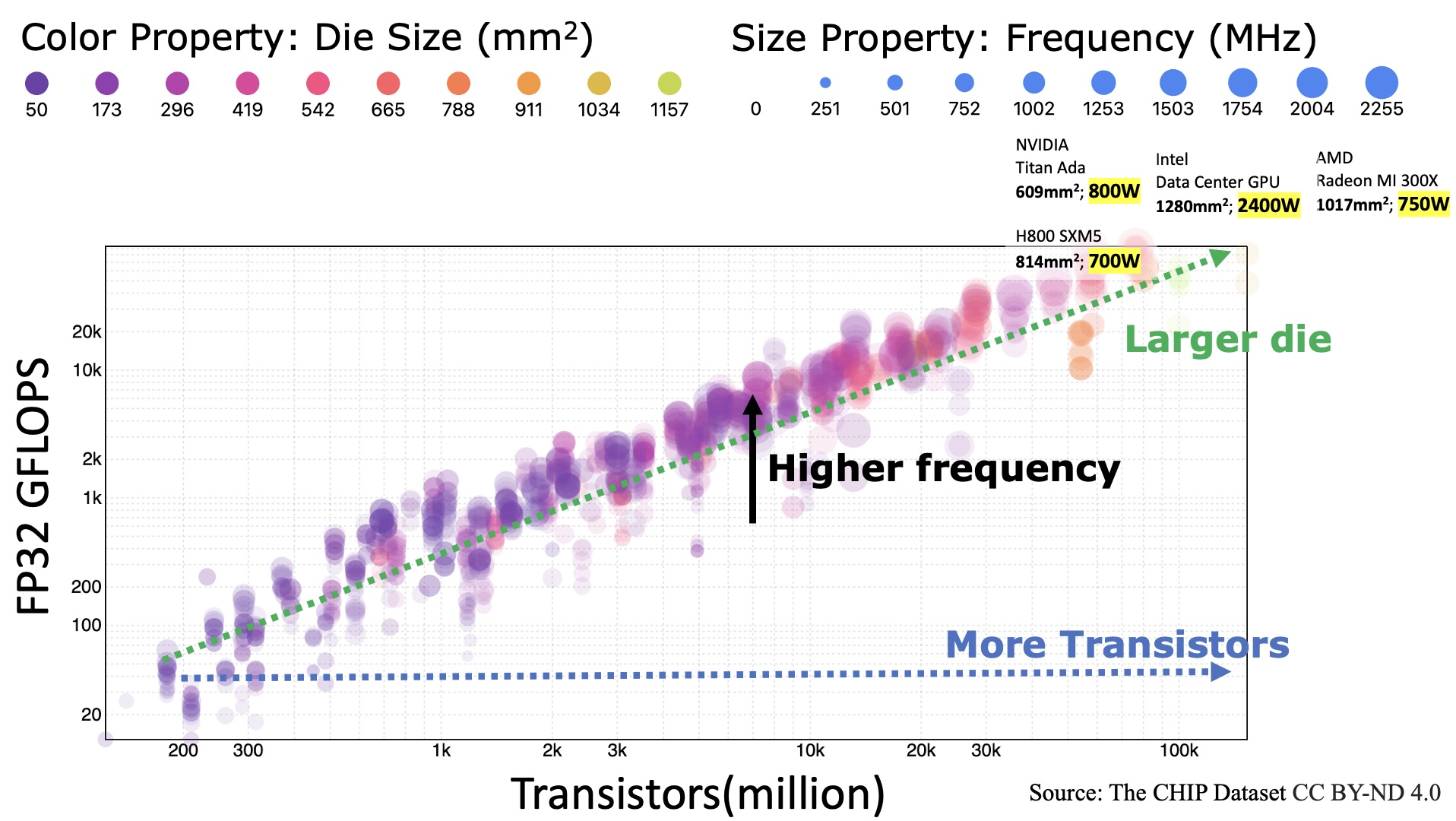}
  \caption{Significant GPU performance growth from higher transistor density, higher frequency, and larger die size. }
  \label{fig:gpu-scaling}
\end{figure}

Large-scale computing infrastructures today are extremely efficient~\cite{datacenter-facebook-pue,datacenter-google-pue}, with Power-Usage-Effectiveness (PUE) of roughly 1.1.
Performance-per-watt energy efficiency of microprocessors are also steadily improving. For GPUs, the theoretical GFLOPS performance-per-watt doubles every 3 to 4 years~\cite{cpu-gpu-ppw}. Figure~\ref{fig:gpu-scaling} illustrates the notable GPU performance improvement (in FP32 GFLOPS) as a combined effect of higher transistor density, higher frequency, and larger die size. Decades of efficiency optimization across the various dimensions of computer systems has led to orders-of-magnitude energy efficiency improvement for computing. 

Thus, the global data center energy use increased by an estimated 6\% between 2010 and 2018~\cite{Masanet}, despite a 550\% increase in the global datacenter compute instances. 
Thanks to Barroso's foundational contributions in energy proportional datacenter computing~\cite{Barroso-energy-proportionality}, the tail at scale~\cite{Barroso-tail-at-scale}, warehouse-scale computing infrastructures have become extremely energy efficient and significantly more cost-effective today~\cite{Barroso-WSC}. 
This, in turn, influenced the 
seismic shift for computation demand from personal computers and traditional datacenters to warehouse-scale computing infrastructures.

\mbox{}
\vfill
\hrulefill
\\
This is a pre-print of the article that has been accepted and to appear at the IEEE Micro Special Issue on \textit{The Past, Present, and Future of Warehouse-Scale Computing}. 
It is based on the industry experience and key lessons learned from the Green AI journey that many colleagues at Meta have contributed to. We pinpoint new opportunities to sustainably scale AI computing beyond the decades of focus on  efficiency for computing.


Despite the significant advancement in efficiency and cost, as digital technologies become an essential part of humanity, their growing prominence is reflected in computing’s energy footprint. 
Between 2017 and 2021, the electricity consumption of Google, Meta, and Microsoft grew by more than 2-3 times --- AI is the most important application driver to computing's growth at-scale. The International Energy Agency (IEA) projected the data center electricity use to double from the energy consumption of 460 TWh in 2022 to more than 1,000 TWh by 2026~\cite{iea-2024}. 

Workloads shape computing infrastructure design.
New application drivers, such as artificial intelligence (AI), are introducing an enormous change to the overall computing industry --- \textit{at a rate that has not been seen in the history of computing before}.  
The rise of Generative AI models, such as, ChatGPT~\cite{chatgpt}, LLaMA~\cite{touvron2023llama}, DALL-E~\cite{dall-e}, Sora~\cite{videoworldsimulators2024}, have demonstrated impressive results on generation tasks for a wide range of modalities, in the form of images, videos, language text, and speech. Today’s ChatBots and AI assistants are only scratching the surface of generative AI technologies as we consider personalized AI assistants that learn and recommend videos to users, write their own articles, solve math problems, create novel music and arts. While unlocking the potential of fueling significant economic growth and boosting productivity, scaling modern AI technologies to billions of people demands a prohibitive amount of computing capabilities, energy, and environmental resources. 

AI's system infrastructure requirements are paramount, requiring a few to hundreds of thousands of training accelerators to develop a foundation model, and many more for inference for real-time millisecond query-serving deployment. 
Depending on the machine learning tasks, the number of GPUs used per model can vary dramatically. For example, training a state-of-the-art text-to-image/video generation model may use \textit{14 times more} GPUs per model parameter than that of large language models (LLMs) for industry-scale use cases~\cite{golden2023generative}. \textit{Optimizing efficiency across the entire deep learning model development cycle} is particularly important now to scale modern AI in a cost-effective manner and to sustainably accelerate the realization of artificial general intelligence.

\begin{figure}
\centering
\begin{subfigure}[b]{0.45\textwidth}
   \includegraphics[width=1\linewidth]{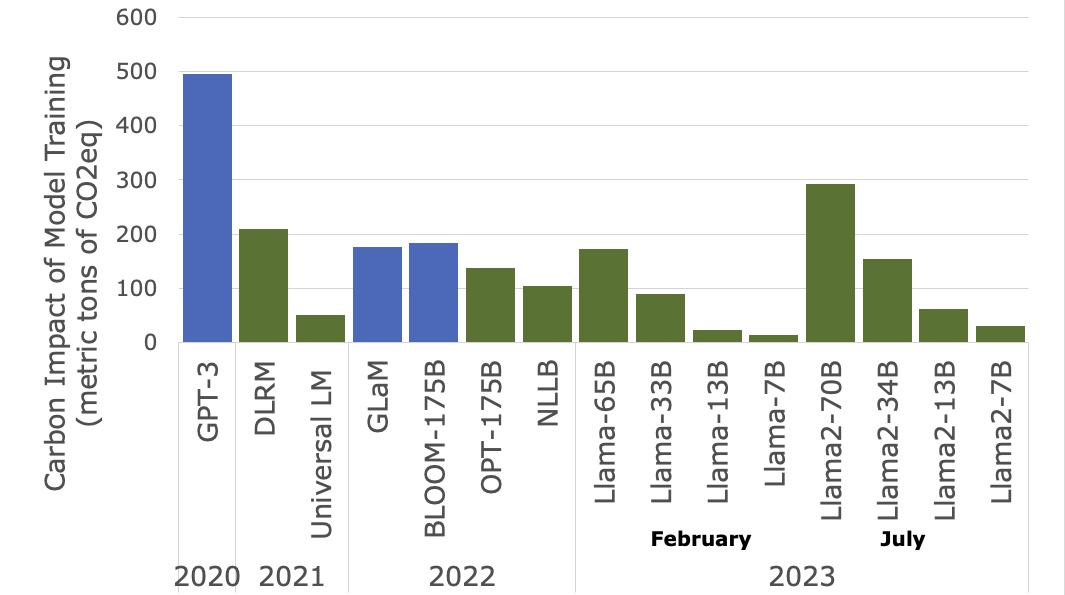}
   \caption{}
   \label{fig:carbonimpact-ai-ops-all-v2} 
\end{subfigure}

\begin{subfigure}[b]{0.4\textwidth}
   \includegraphics[width=1\linewidth]{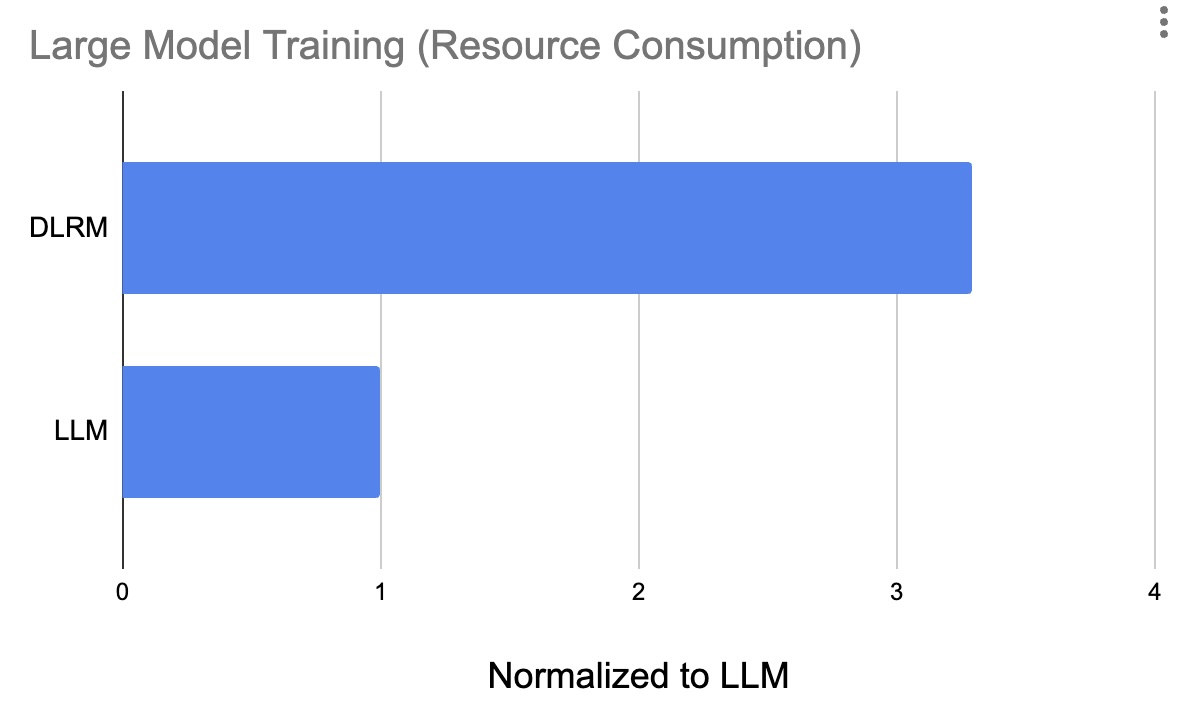}
   \caption{}
   \label{fig:dlrm-llm-production}
\end{subfigure}

\caption{(a) \textit{Operational} carbon impact of open-source AI model training: GPT-3~\cite{brown2020language}, DLRM and Universal LM~\cite{wu2022sustainable}, GLaM, BLOOM-175B~\cite{du2022glam}, NLLB~\cite{nllb}, Llama-65/33/13/7B~\cite{touvron2023llama}. Open-source models from Meta are marked in green. The carbon impact result assumes fixed carbon intensity for electricity. When considering location-based carbon intensity of electricity, the carbon impact of BLOOM-175B is 30 tonnes CO2eq. In addition, corporate-level sustainability programs mitigate model training's carbon emissions. (b) Large model training resource consumption (normalized to Large Language Model (LLM)): Deep learning recommender systems require more than three times higher training resources than LLM training.}
\end{figure}

\section{Understanding the Carbon Impact of AI}

To scale AI sustainably, we must understand the carbon impact of AI quantitatively across its lifecycle. AI's overall lifecycle carbon footprint comes from \textit{manufacturing} --- 
carbon emissions from manufacturing infrastructures and hardware specifically for AI (\underline{embodied carbon footprint}) and 
\textit{product use} --- carbon emissions from the use of AI (\underline{operational carbon footprint}).

\noindent \textbf{Operational Carbon Footprint: Training} Figure~\ref{fig:carbonimpact-ai-ops-all-v2} presents the operational carbon footprint of model training for key open-source AI technologies: GPT-3~\cite{brown2020language}, DLRM and Universal LM~\cite{wu2022sustainable}, GLaM, BLOOM-175B~\cite{du2022glam}, NLLB~\cite{nllb}, Llama-65/33/13/7B~\cite{touvron2023llama}. 
Training a universal language translation model (LM) produces 45.2 metric tons of carbon emissions whereas training a state-of-the-art deep learning recommendation model (DLRM), on average, produces over 4 times higher carbon emissions than that of the language model~\cite{wu2022sustainable}. 
{\camerareadyaddition We continue to observe that newer generation machine learning models come with better capabilities in model quality. For example, Meta's Llama3-70B model -- pre-trained and instruction tuned generative text model, comes with more than 6 times higher carbon footprint as compared to Llama2-70B. The higher training carbon footprint is due to a larger training corpus and an increased context window size, resulting in higher machine learning capabilities. However, it is also important to note that the training carbon footprint of open-source models, such as the family of Llama models, can be more effectively amortized over all use cases worldwide, significantly reducing the energy demand and environmental footprint from individual entities repeating the development of their own models.}
In addition to the open-source AI models, Figure~\ref{fig:dlrm-llm-production} presents the cumulative resource usage for training the top three DLRM tasks, normalized to that of LLM. DLRM tasks demand upto 3 times more training resources than the LLM. 

\noindent \textbf{Operational Carbon Footprint: Inference} Once the model is trained, it is further optimized for deployment, for which additional carbon footprint is incurred. 
When considering inference, the operational carbon footprint of AI across the model development cycle can be increased by 2-3 times for the DLRM and LM tasks~\cite{wu2022sustainable}. The exact ratio depends on \textit{the need and frequency of new model development}.

\begin{figure}
\centering
\begin{subfigure}[b]{0.45\textwidth}
   \includegraphics[width=1\linewidth]{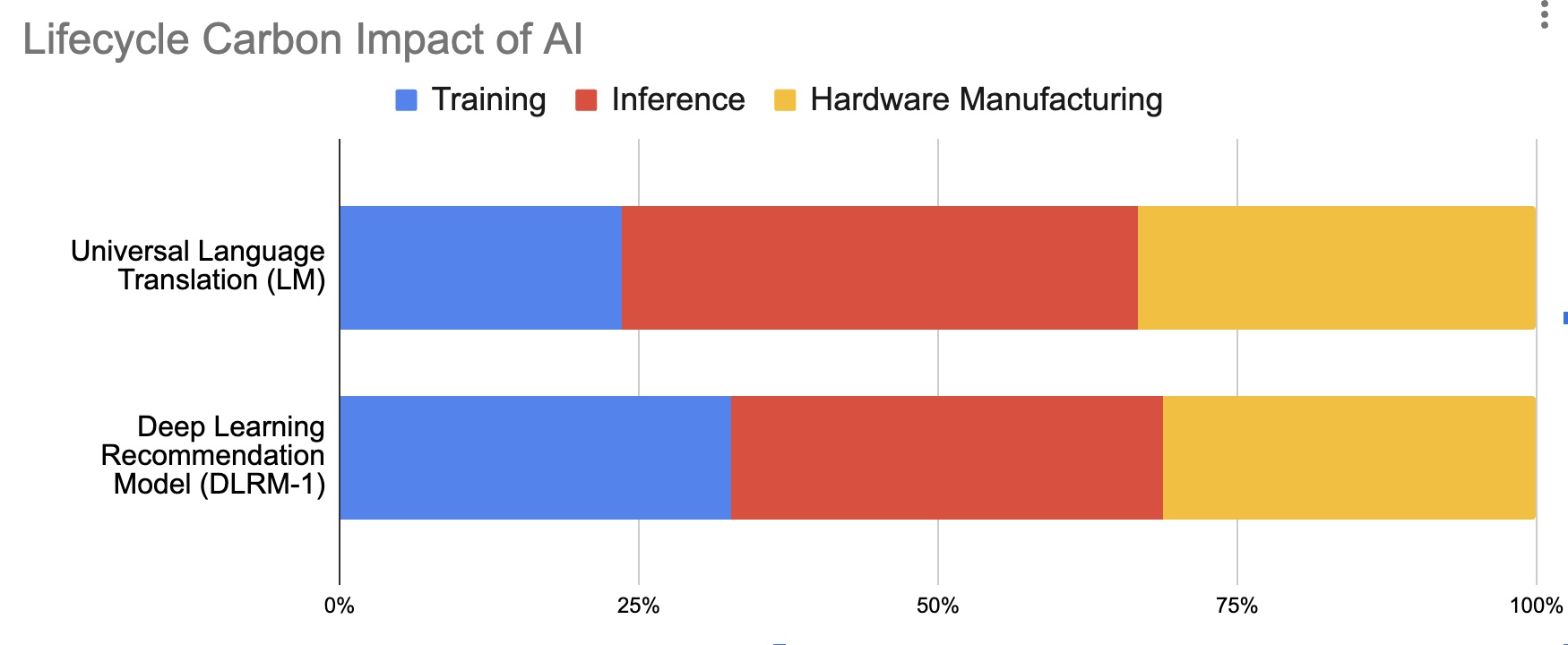}
   
   \caption{}
   \label{fig:carbonimpact-ai-cap} 
\end{subfigure}

\begin{subfigure}[b]{0.425\textwidth}
   \includegraphics[width=1\linewidth]{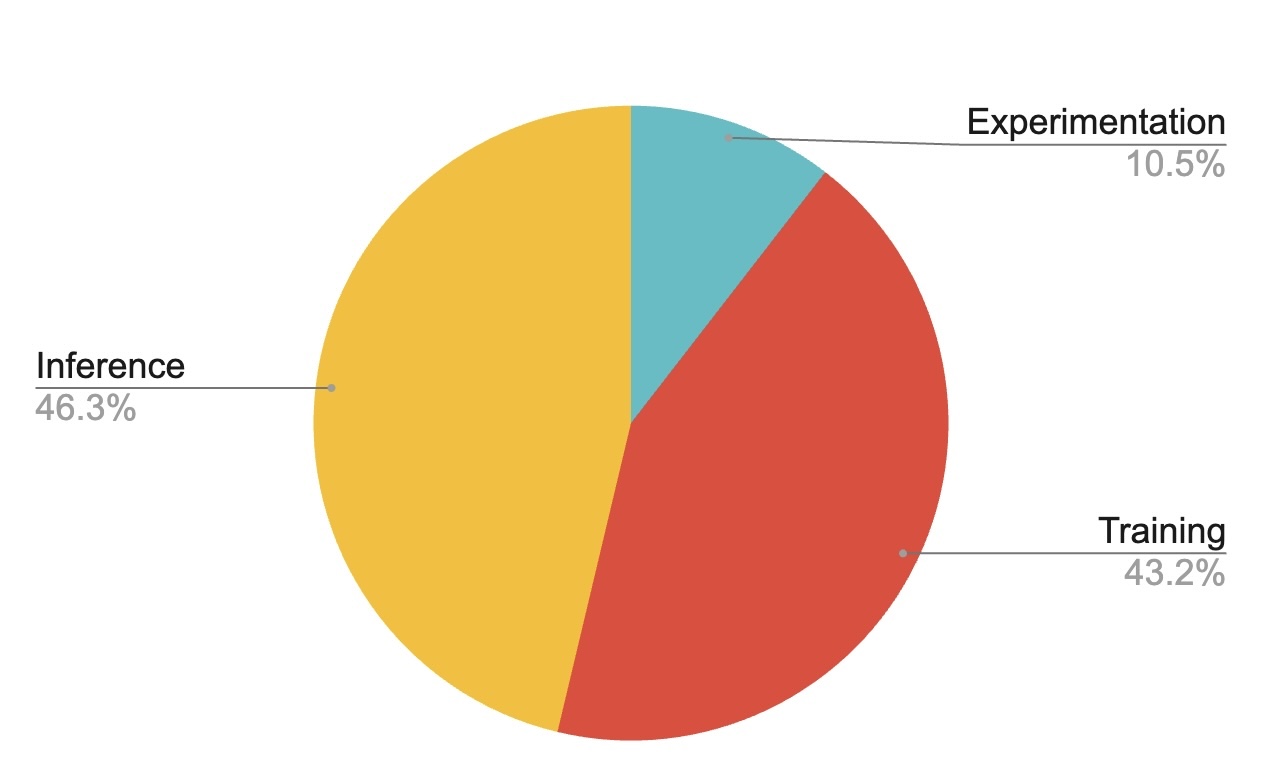}
   \caption{}
   \label{fig:fleet-view}
\end{subfigure}

\caption{(a) \textit{Lifecycle} carbon impact of the Multi-Lingual Language Translation model (Universal LM in Figure~\ref{fig:carbonimpact-ai-ops-all-v2}) and the Deep Learning Recommendation Model (DLRM-1). Over the model development lifecycle, the operational carbon impact of model training vs. inference is about 1 to 2 for Universal LM and 1 to 1 for DLRM-1. When considering the overall AI system lifecycle, embodied carbon footprint from hardware manufacturing introduces an additional 50\% of the operational carbon emissions. Reducing any part of the lifecycle emissions -- Training vs. Inference; Operational vs. Embodied Carbon Emissions -- will translate to lower lifecycle emissions for AI. (b) The power capacity breakdown for the three key phases of the AI development lifecycle -- Experimentation, Training, and Inference.}
\end{figure}

\noindent \textbf{Embodied Carbon Footprint} AI’s carbon footprint goes beyond operational energy use. Systems used for model training and deployment come with manufacturing carbon emissions that are embedded in the hardware used for AI. 
Figure~\ref{fig:carbonimpact-ai-cap} illustrates the carbon footprint breakdown for the multi-lingual language translation model (University LM in Figure~\ref{fig:carbonimpact-ai-ops-all-v2}) --- operational to embodied carbon footprint is approximately 2 to 1. Optimizing operational and/or embodied carbon footprint can both translate to improvement in the overall \textit{lifecycle carbon impact of AI}.

Embodied carbon in the hardware has become one of the most significant source of carbon footprint for warehouse-scale computing infrastructures~\cite{meta-sustainability-2023,chasingcarbon}. Datacenter construction and hardware manufacturing carbon footprint has increased from 41\% to over 60\% of Meta's Scope 3 Greenhouse Gas (GHG) emissions between 2019 and 2022. The Scope 3 GHG emission category is similarly a significant factor for Google's computing infrastructures~\cite{google-sustainability-2023}. 

For battery-operated consumer electronics, the stringent requirement of low power and low latency for smartphones means significant effort has been put into optimizing for operational efficiency.
This has led to an ever-increasing degree of accelerators -- general-purpose programmable as well as domain-specific ones -- for smartphone hardware~\cite{hill2021acceleratorlevel}. Using advanced semiconductor manufacturing technologies further improves the operational efficiency. However, more accelerators translate into larger dies; more advanced semiconductor technologies translate into higher manufacturing GHG emissions~\cite{imec-iedm}.
Embodied carbon footprint is the dominating source of the computing system's overall lifecycle emissions. 
Taking iPhone 15 Pro released in 2023 as an example, embodied carbon footprint is 83\% of the product lifeycycle emissions whereas operational carbon footprint is only 15\% with the remaining 2\% emissions from transportation and end-of-life processing~\cite{apple-iphone15-sustainability}. This carbon footprint breakdown highlights the hidden cost of dark silicon -- embodied carbon emissions.

\section{Efficiency and Beyond}

To reduce the significant carbon impact of AI, we have a lot to learn from Barroso's foundational work in energy proportionality for computing. To bend the ever-increasing resource demand of AI, we must accelerate efficiency optimization across all layers of the AI system stack. {\camerareadyaddition At the same time, some key differences exist when strictly optimizing for energy efficiency or TCO versus optimizing for carbon footprints.}

In terms of where carbon and TCO align, hardware-software co-design and optimization can effectively reduce AI's operational energy and carbon footprint. Taking the multi-lingual language model as an example, a combination of data locality optimization, GPU acceleration, low precision data format use, and algorithmic optimization can bring \textit{over 800 times} energy efficiency improvement~\cite{wu2022sustainable}. Efficiency optimization across the following key dimensions leads to multiplicative improvement on the operational carbon impact of AI.

\begin{itemize}

   \item \textbf{Data}: AI models train on massive amounts of data. Model performance increases with data scale, following a power law scaling \cite{kaplan2020scaling}. 
   Recent works have shown that power law scaling can be significantly improved if data is selected for model training. Furthermore, the benefits of data pruning grows as data sizes increase\cite{sorscher2022beyond, sachdeva2021svpcf}.
   When designed well, data scaling, sampling and curation strategies can result in substantial hardware resource efficiency improvement while achieving faster training time and higher model quality. Complementary to AI data optimization, the data storage and ingestion pipeline for AI demands significant power capacity~\cite{zhao-isca2022}. An optimized composite data storage infrastructure using novel application-aware cache policies can absorb more than 3 times higher IO than a baseline LRU flash cache, reducing power demand in a petabyte-scale production AI training cluster by 29\%~\cite{zhao-atc2023}.

\begin{figure}
\centering
\begin{subfigure}[b]{0.45\textwidth}
   \includegraphics[width=1\linewidth]{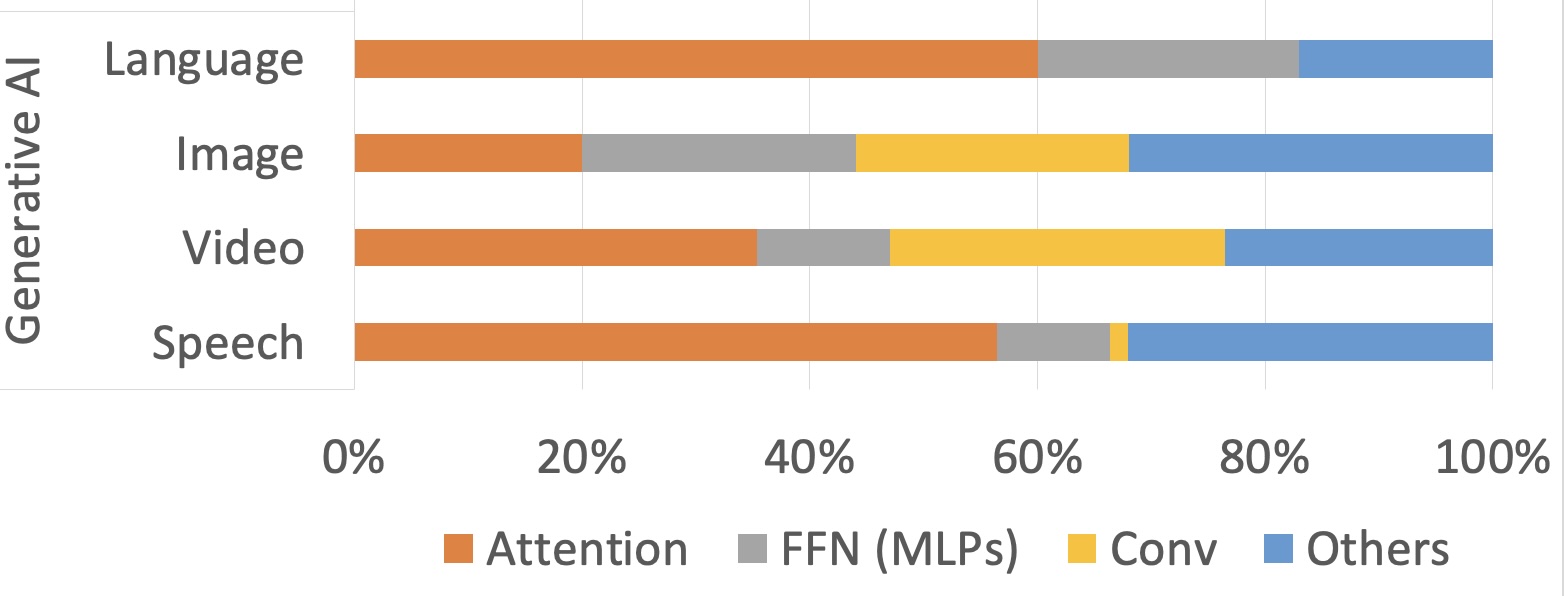}
   \caption{}
   \label{fig:carbonimpact-ai-operatortime} 
\end{subfigure}

\begin{subfigure}[b]{0.4\textwidth}
   \includegraphics[width=1\linewidth]{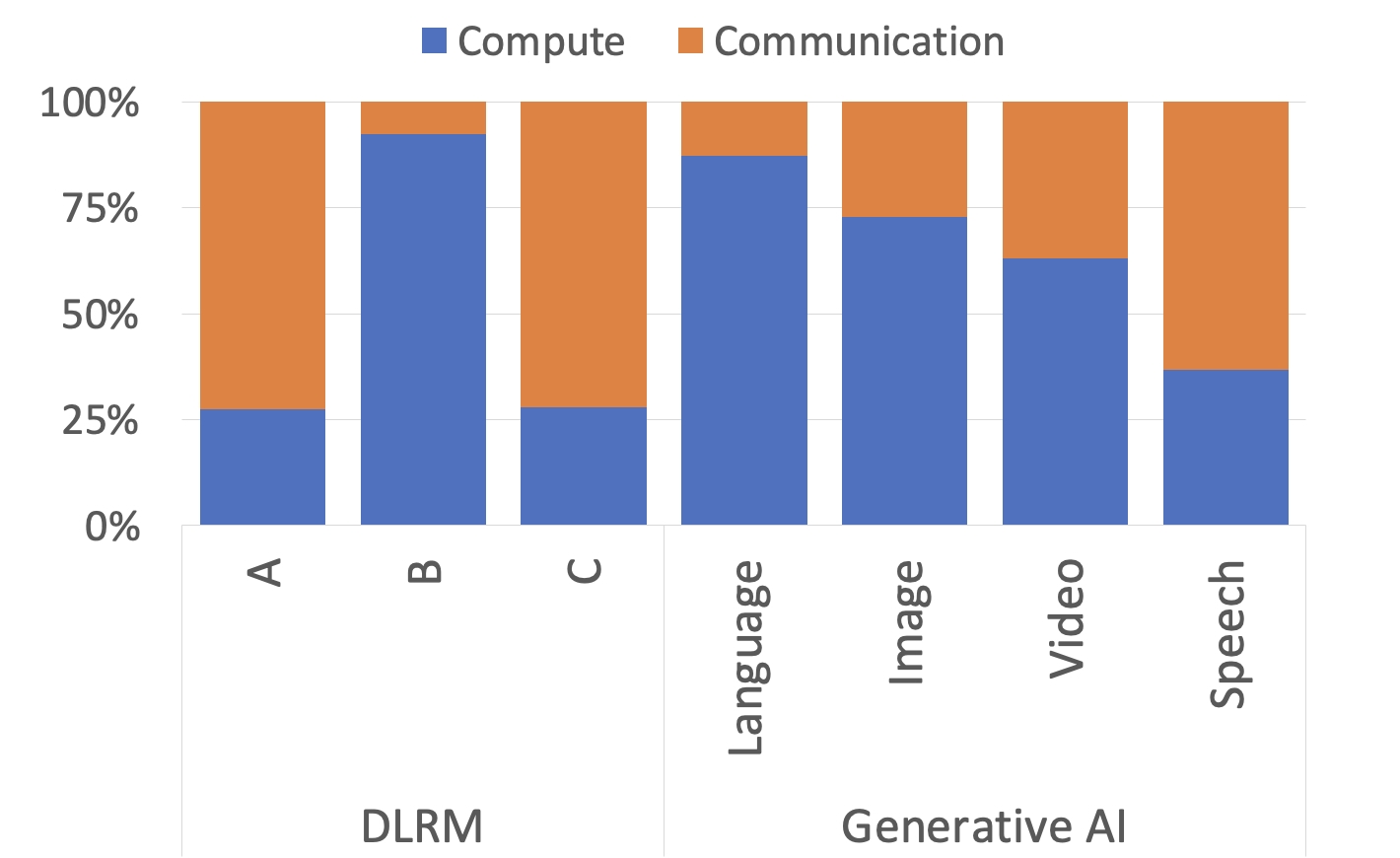}
   \caption{}
   \label{fig:carbonimpact-ai-compute-communication}
\end{subfigure}

\caption{(a) Across the key generative AI tasks, over language, image, video, and speech tasks, Attention, Feedforward Network (FFN), and Convolution operations are the targets for timing performance optimization. As compared to deep learning recommendation models, that contribute to 79\% of AI inference cycles at Meta's datacenter fleet in 2018~\cite{gupta2020hpca}, we expect a shift in \textit{where time is spent} from \textit{embedding} to the other key operators as generative AI tasks become increasingly deployed at-scale. (b) Communication contributes significantly to the overall model training time performance. Optimizing Communication is increasingly important as Compute is improved through model-hardware co-design.}
\end{figure}

    \item \textbf{Models}:
    Parameter-efficient machine learning models can contribute to significant carbon footprint reduction. Taking the family of foundational language models, Llama, as an example~\cite{touvron2023llama}, LlaMA-13B outperforms GPT-3 (175B) for a variety of tasks and, at the same time, consumes approximately ~\textit{24 times lower energy than GPT-3}. Llama as a parameter-efficient model for language tasks is superior across the key design dimensions of accuracy, training time, energy and carbon footprint.
    
    \item \textbf{Systems}: 
    Systems designed to efficiently accelerate common computation primitives for AI further reduces operational energy and carbon footprint. Figure~\ref{fig:carbonimpact-ai-operatortime} illustrates that the Attention module~\cite{attention} is a common operator across the generative AI tasks, from language~\cite{touvron2023llama} to image, video, and speech tasks~\cite{seamless,golden2023generative}. Accelerating the execution of \textit{Attention} via new system innovations, such as FlashAttention~\cite{dao2022flashattention}, can translate into substantial latency and energy footprint reduction. 
    
    As the performance of GPUs continues to improve, the overall model execution time will increasingly depend on the effectiveness of the communication collectives and the underlying networking technologies. 
    Figure~\ref{fig:carbonimpact-ai-compute-communication} illustrates that, across Deep Learning Recommendation Models (DLRMs), LLMs, and Multi-Modal (MM) AI technologies, communication optimization can also lead to substantial training throughput improvement~\cite{hsia2023mad}. 
    However, we must balance the operational efficiency improvement with embodied carbon stemmed from \textit{newer generation of hardware} to optimize for \textit{total lifecycle emissions}. 
    
    \item \textbf{Datacenters}: As machine learning model training scales out to tens of thousands of AI accelerators housed in warehouse-scale computing infrastructures, it can benefit from hyperscale datacenters that leverages economy of scale. For example, more efficient cooling and power delivery solutions can be deployed at the datacenter scale, shared by all hardware in the same datacenter building, as opposed to using server- or blade-level cooling. As model training continues to scale out, we must design datacenters with reliability and fault-tolerance in mind. Failure in hardware, let it be from GPUs, memory, storage, or network equipment, will directly impact the lifecycle emissions of AI. In particular, failure probability can increase exponentially as the scale of model training increases. While reliability through hardware redundancy comes with additional embodied carbon, the lifecycle emissions can be a net win when factoring in operational efficiency improvement.

    \item \textbf{Energy}: Minimizing the carbon intensity of electricity powering computing infrastructures is another lever to reduce the carbon impact of AI. When spare computing capacity is available, mapping computations, such as model training or AI-powered services, to where renewable energy is abundant can lower the environmental impact. However, it comes with unique design challenges. For model training, due to data protection regulations, AI datasets may not always be available in datacenters located in different geographic regions. And, for inference with real-time limitation, product quality may be impacted. Ultimately, if electricity from the power grid is clean, it benefits datacenters and all other consumers. It must be realized with minimal manufacturing carbon and in a cost-effective way~\cite{carbon-explorer}. 
\end{itemize}

{\camerareadyaddition That said, there are notable instances where TCO-neutral or TCO-negative design decisions are actually positive from a carbon perspective, such as the choice of whether and how much to leverage clean energy, battery storage, time-shifting compute, or other tactics. Therefore, expanding our approach to balance these competing interests is key.}

\textbf{Sustainability} means "\textit{[d]evelopment that meets the needs of the present without compromising the ability of future generations to meet their own needs}" from \textit{Our Common Future} by the United Nation in 1987.
Despite the importance for optimizing energy efficiency across AI model development and hardware life cycles, 
AI's resource demand is growing rapidly. 
While algorithmic efficiency and domain-specific hardware systems can improve the \textit{operational energy footprint} of AI model training by more than 90\%~\cite{wu2022sustainable,patterson2021carbon}, efficiency improvement has encouraged higher uses, leading to ever-increasing computing resource consumption. This is a phenomenon that is known as the Jevon's paradox. 
Thus, to scale AI technologies sustainably, we must understand \textit{the rebound effect}'s implications on the power grid infrastructure and on the environment.

\begin{itemize}
    \item \textbf{Power capacity is the limiting factor at-scale.}  The rapid growth in computing is putting significant stress on power grids. Power grids, such as EirGrid in Ireland, may not have energy production capacity to source large power loads of datacenter computing~\cite{ireland-dc-1}. Datacenters already used around 4,000 GWh of electricity in Ireland, corresponding to 14\% of Ireland’s total electricity used in 2021~\cite{ireland-dc-3} and the datacenter electricity need is likely to increase to 30\% of Ireland's annual electricity supply by 2029~\cite{ireland-dc-2}. For other power grids that are able to support the large power capacity requirement of datacenters, making it cost-effective and environmentally-sustainable is a key challenge. In northern Virginia, where the largest concentration of datacenters in the world is at, Dominion Energy -- the Virginia Electric and Power Company -- considered grid infrastructure designs that balance between cost and carbon emissions~\cite{virginia-dc-2}. It does so by considering a wide variety of energy generation options, from building renewable energy and storage for the power grid to extending the life of coal and natural gas energy generation infrastructure while also considering small modular nuclear reactors~\cite{virginia-dc-1}.
    \item \textbf{Embodied carbon is becoming the dominating source of the lifecycle carbon footprint for computing.} From iPhone 3 (2008) to iPhone 15 (2023), the lifecycle carbon emissions due to hardware manufacturing has increased from 49\% to 83\%. The operational carbon footprint reduced by 2.39x while manufacturing carbon footprint increased by 2.2x. Reducing embodied carbon stemmed from more complex hardware design and advanced semiconductor manufacturing is key to sustainable computer systems~\cite{sigarch-embodied-carbon}. Developing expandable hardware and software stack that facilitate significantly longer lifetimes helps mitigate embodied carbon as well. 
    \\
    However, it is extremely challenging to optimize for lifecycle carbon emissions for AI in the presence of fast-evolving algorithmic innovations. Even though application-specific hardware comes with significant operational efficiency improvement, the operational carbon footprint improvement can be over-shadowed by the additional manufacturing, embodied carbon emissions --- an optimal hardware refreshment period that minimizes the lifecycle carbon emissions depends on system hardware design and operational efficiency benefits. 
    
    \item \textbf{Hundreds of millions of servers and other hardware IT equipment in the cloud and billions of consumer electronics in the world will reach end-of-life in less than 10 years, with an average of 3-4 years for consumer electronics.}
    Designing computer systems with modularity and right-to-repair in mind can lead to effective e-waste reduction and upcycling opportunities. Modular systems enable component-level upgrades without having to decommission the system at its entirety, reducing overall planetary impact. Designing systems with repairability in mind increases product lifetime, leading to better amortization of embodied carbon footprint. Finally, mining raw materials and metals, such as copper, lithium, silver, gold, nickel, and aluminum, for electronics produces significant GHG emissions. Extracting precious metals, such as gold, from old electronics circuit boards is already demonstrated with significant business potential~\cite{ewaste-mint} while upcycling aluminum for, e.g., future iPhones, reduces the product's overall lifecycle emissions as well as the amount of e-waste from consumer electronics.
\end{itemize}

\section{Looking Forward}

Understanding state-of-the-art carbon emission quantification methodologies using the Greenhouse Gas (GHG) Protocol~\cite{ghg} and Life Cycle Assessment (LCA)~\cite{lca} is an important first step. The GHG protocol defines an accounting standard for carbon emissions and equivalent (CO2e) of a company. There are three GHG categories: Scope 1 (direct emissions), Scope 2 (indirect emissions from purchased energy), and Scope 3 (indirect emissions from upstream/downstream uses). 

\begin{figure}
\centering
\begin{subfigure}[b]{0.45\textwidth}
   \includegraphics[width=1\linewidth]{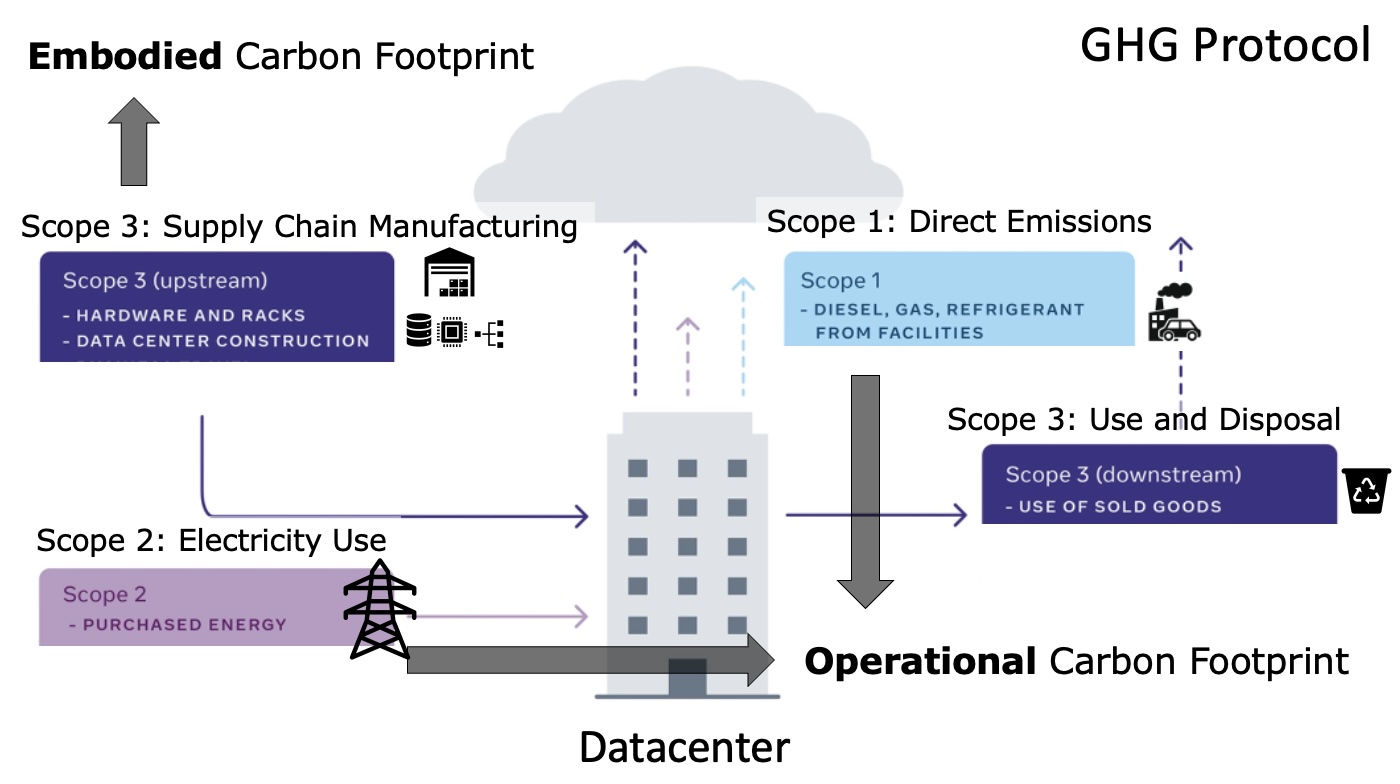}
   \caption{}
   \label{fig:ghg-dc} 
\end{subfigure}

\begin{subfigure}[b]{0.45\textwidth}
   \includegraphics[width=1\linewidth]{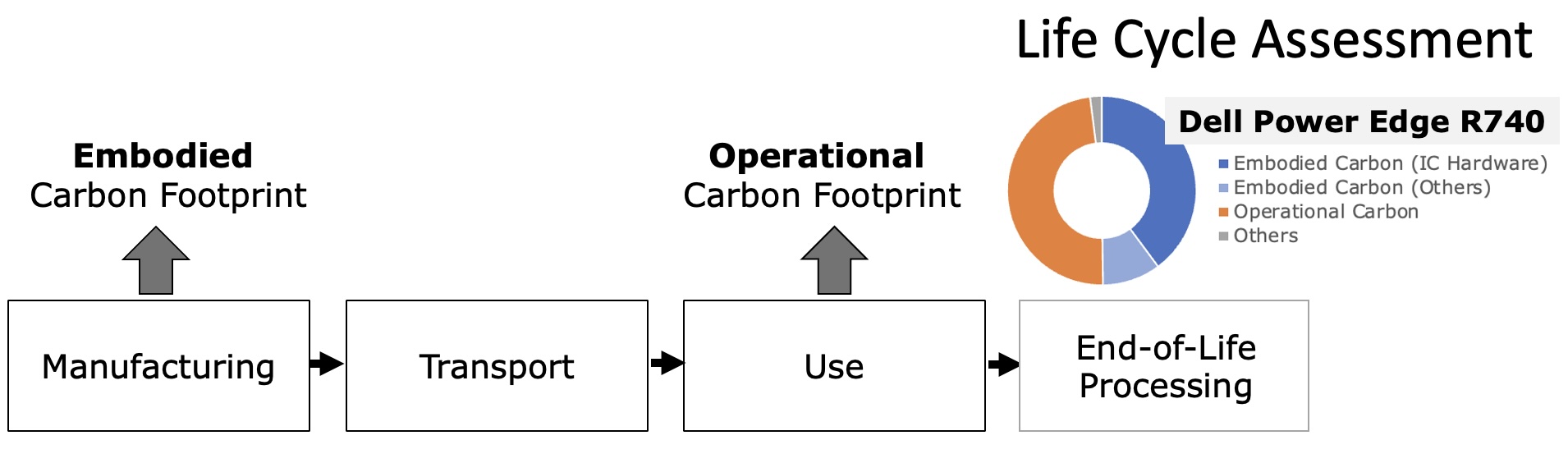}
   \caption{}
   \label{fig:lca-dc}
\end{subfigure}

\caption{(a) Computing's carbon emissions for datacenter computing using the Greenhouse Gas (GHG) Protocol. (b) Computing's carbon emissions for datacenter hardware using Life Cycle Assessment (LCA).}
\end{figure}

From the perspective of a datacenter operator, Scope 1 emissions come from fuel combustion in offices and datacenters, Scope 2 come from purchased energy produced at respective power grids, and Scope 3 come from all other activities, such as IT hardware manufacturing, datacenter construction, and others (Figure~\ref{fig:ghg-dc}). Here, the operational footprint of a datacenter is the sum of Scope 1 and Scope 2 emissions of the datacenter whereas the embodied carbon footprint is the part of the Scope 3 emissions from datacenter construction and hardware manufacturing. 
Based on the publicly-available sustainability reports using the GHG protocol, in 2022, 67.9\%, 47.4\%, 40.1\% 51.7\% of datacenter carbon emissions by Meta~\cite{meta-sustainability-2023}, Google~\cite{google-sustainability-2023}, Microsoft~\cite{microsoft-sustainability-2023}, and Oracle~\cite{oracle-sustainability-2024} come from datacenter construction and hardware manufacturing. Here, embodied carbon emissions account for the emissions from Capital Goods, Upstream Transportation and Distribution, Purchased Goods and Services.

In addition to organization-level analysis using the GHG protocol, LCA is a standard practice used to evaluate environmental impact of individual hardware products. There are four major phases over a product's life: Manufacturing, Transport, Use, and End-of-life Processing (Figure~\ref{fig:lca-dc}). 
Using the LCA methodology, the operational footprint of a computer is the product of its energy consumption and carbon intensity of electricity during the use phase whereas the embodied carbon footprint is the carbon footprint to procure raw materials, manufacture wafers, fabricate integrated circuits, packaging and assembling the system. Based on publicly-available LCA reports, 49.8\% of the lifecycle emissions of Dell PowerEdge R740, a general-purpose rack server, comes from hardware manufacturing~\cite{dell-r740}, 80\% of which is due to IC manufacturing. In comparison, 83\% of the lifecycle emissions of iPhone 15 Pro (128GB) comes from hardware manufacturing with 15\% from product use and the remaining 2\% from transportation and end-of-life processing. 
Having a framework for sustainability in computing elucidates the problem space.

Taking a step beyond the first principle analysis, we need more granular, higher quality carbon telemetry, datasets, metrics, in order to enable sustainability as a computer system design principle:

\noindent \textbf{Carbon Telemetry:} \textit{We cannot reduce what cannot be measured.} 
We need high fidelity tools to measure the lifecycle emissions of a computer system -- both operational \textit{and} embodied carbon footprint. 
However, it is challenging to do so systematically today. Characterizing and analyzing carbon emissions is a complex process, as compared to performance measurement, power and energy modeling. 

To enable and accelerate environmentally sustainable computing,
recent research studies explored and built carbon modeling frameworks, such as ACT~\cite{act}, Carbon Explorer~\cite{carbon-explorer}, GreenChip~\cite{GreenChip}, FOCAL~\cite{GreenChip}.
Expanding upon the early carbon modeling research frameworks, imec.netzero (~\url{https://netzero.imec-int.com/}) is developing an industry-grade carbon quantification framework to further advance semiconductor manufacturing and integrated circuit design by enabling carbon emissions and equivalent as a first-class design principle along performance, power, and cost.

\noindent \textbf{Carbon Dataset:}
At the datacenter fleet level, Meta started developing and scaling a first-of-its-kind carbon dataset based on the best available embodied carbon estimates at
the scale of the hundreds of millions of components in Meta's
data center hardware~\cite{meta-sustainability-2023}. This dataset lays the foundation for embodied carbon reductions by enabling Meta to make a data-driven decision for lowering value chain carbon emissions. At the same time, a carbon dataset that characterizes the manufacturing and operational carbon emissions of AWS EC2 instances is being developed~\cite{aws-ec2-carbon}. Making more carbon datasets for cloud computing helps advance sustainability of the hardware-software ecosystem. 
Furthermore, having the dataset for emissions embodied in the hardware components opens up new opportunities for optimization, such as designing ML model architectures and hardware jointly, to reduce an AI model's overall carbon footprint. Hardware-aware neural architecture search techniques can include embodied carbon costs as part of the multi-objective optimization search process.

\noindent \textbf{Carbon Impact Disclosure:} As AI's demand on computing and energy capacity increases, the impact on the environment can be significant. Measuring and reporting the carbon impact of AI is an important act towards responsible technology development~\cite{henderson2022systematic}. When carbon impact of AI is disclosed in a transparent manner, it drives forward progress. Just within the past three years, many key AI breakthroughs are published with Carbon Impact Statements, i.e., Hugging Face's BLOOM (176B)~\cite{luccioni2022estimating}, Meta's Open Pre-train Transformer (OPT) Language Model~\cite{opt}, Llama2 Open Foundation and Fine-Tuned Chat Models~\cite{touvron2023llama}, Meta's No Language Left Behind (NLLB) Machine Translation Model~\cite{nllb}, Google's GLaM~\cite{patterson2022carbon}. Carbon impact assessment and disclosure, when becoming ingrained in the AI technology development process, drives responsible innovations.

\begin{figure}
  \centering
  \includegraphics[width=0.49\columnwidth]{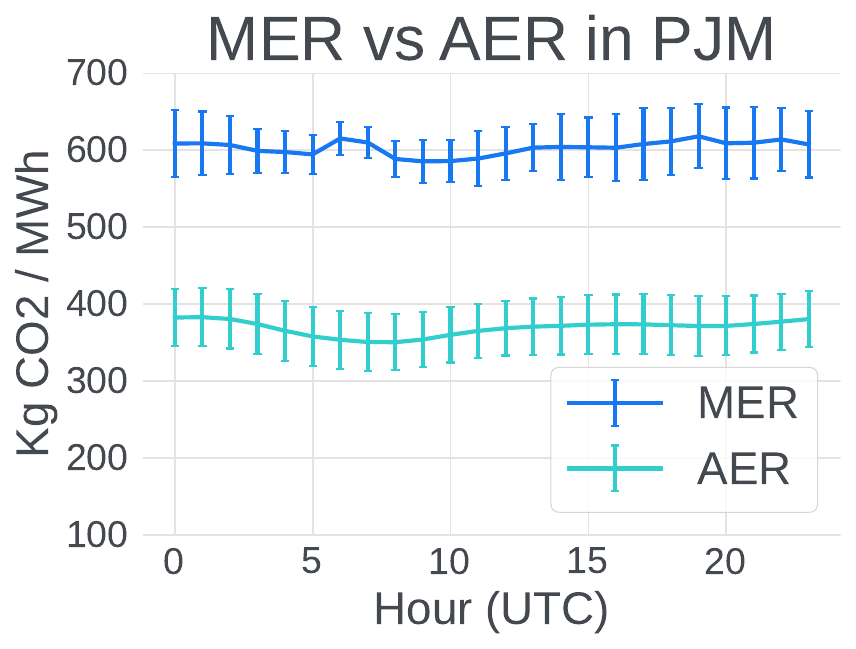}
  \includegraphics[width=0.49\columnwidth]{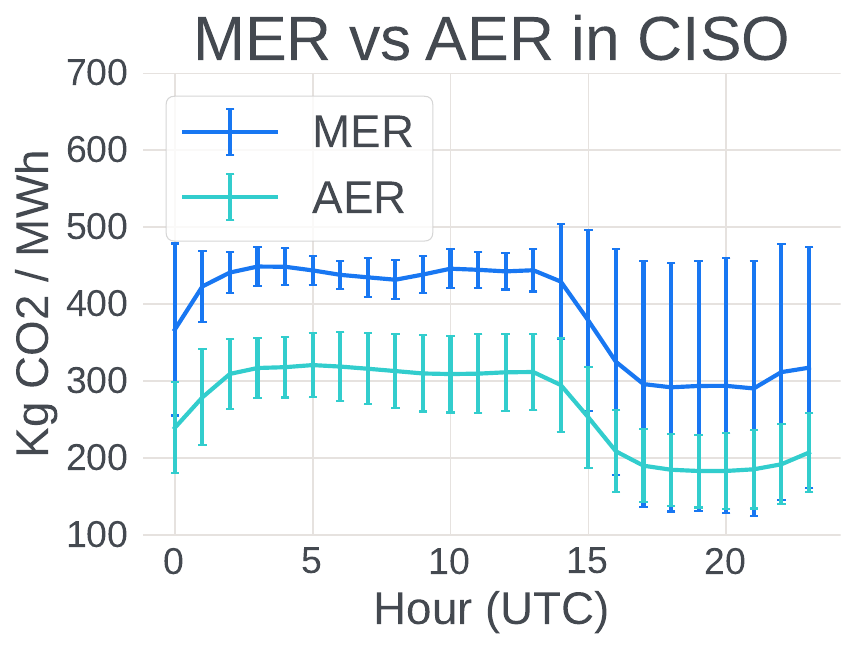}
  \vspace{-2mm}
  \caption{Average and marginal emissions comparison at CISO \& PJM hourly average for year 2022. Error bars represent standard deviation.}
  \label{fig:mer-aer}
\end{figure}

\noindent \textbf{Carbon Metrics:} Many carbon-related metrics have been explored to align sustainability optimization from the perspective of computer system design and management. \textit{CFE} (Carbon Free Energy) is used to guide Google's 24/7 carbon-free datacenter computing whereas \textit{Exergy} is proposed as a measure for the available energy in a system, used to guide the sustainability design space for computer systems~\cite{totally-green}. In addition, new metrics, such as \textit{CDP} (Carbon-Delay Product), \textit{CEP} (Carbon-Energy Product)~\cite{act}, tCDP (total Carbon-Delay Product)~\cite{elgamal2023design}, are proposed to align design and optimization. 
Knowing \textit{what} carbon metrics to use for \textit{when} is still at a nascent stage while it is intuitive that directly optimizing carbon than energy is more effective since it is a direct measure of carbon emission impacts than energy-based metrics, such as megawatt-hour (MWh).

From the perspective of power grids, 
a key aspect contributing to the carbon intensity of energy produced is \textit{what energy source(s)} is on the margin -- the marginal generation units. If wind generators were on the margin, it translates to a lower carbon emission rate. On the other hand, if a coal unit is on the margin, any additional energy demand in the grid comes with higher carbon emissions. The metric of Marginal Emission Rates (MER) is  defined as, given a location, emission rate of the individual marginal unit at that time. To minimize carbon emissions, the power grid must minimize the likelihood to activate energy generation units of high carbon intensity, thus, lowering MER. On the other hand, Average Emission Rates (AER) which simply measures the average emissions of all of the generator units, is more commonly used for carbon optimization. 

Figure ~\ref{fig:mer-aer} shows the difference between MER and AER metrics at two different locations. First, MER is higher in magnitude compared to AER. This is important as it might change the cost/benefit trade-offs in multi objective optimization. Second, standard deviation of MER is larger than AER at CAISO. This is result of the solar energy generation, which is much higher compared to PJM during daytime hours in CAISO. The variability in MER shows the importance of coping with intermittent renewable energy generation, which is not reflected in AER as significantly.

While what metric to use for demand-response scheduling or renewable energy procurement strategies, is still an open question for at-scale deployment, it is clear that coordination between datacenters and the grid is necessary to achieve a clean energy future in a cost-effective and reliable way~\cite{tcr-2023}.

\begin{figure}
  \centering
  \includegraphics[width=0.95\columnwidth]{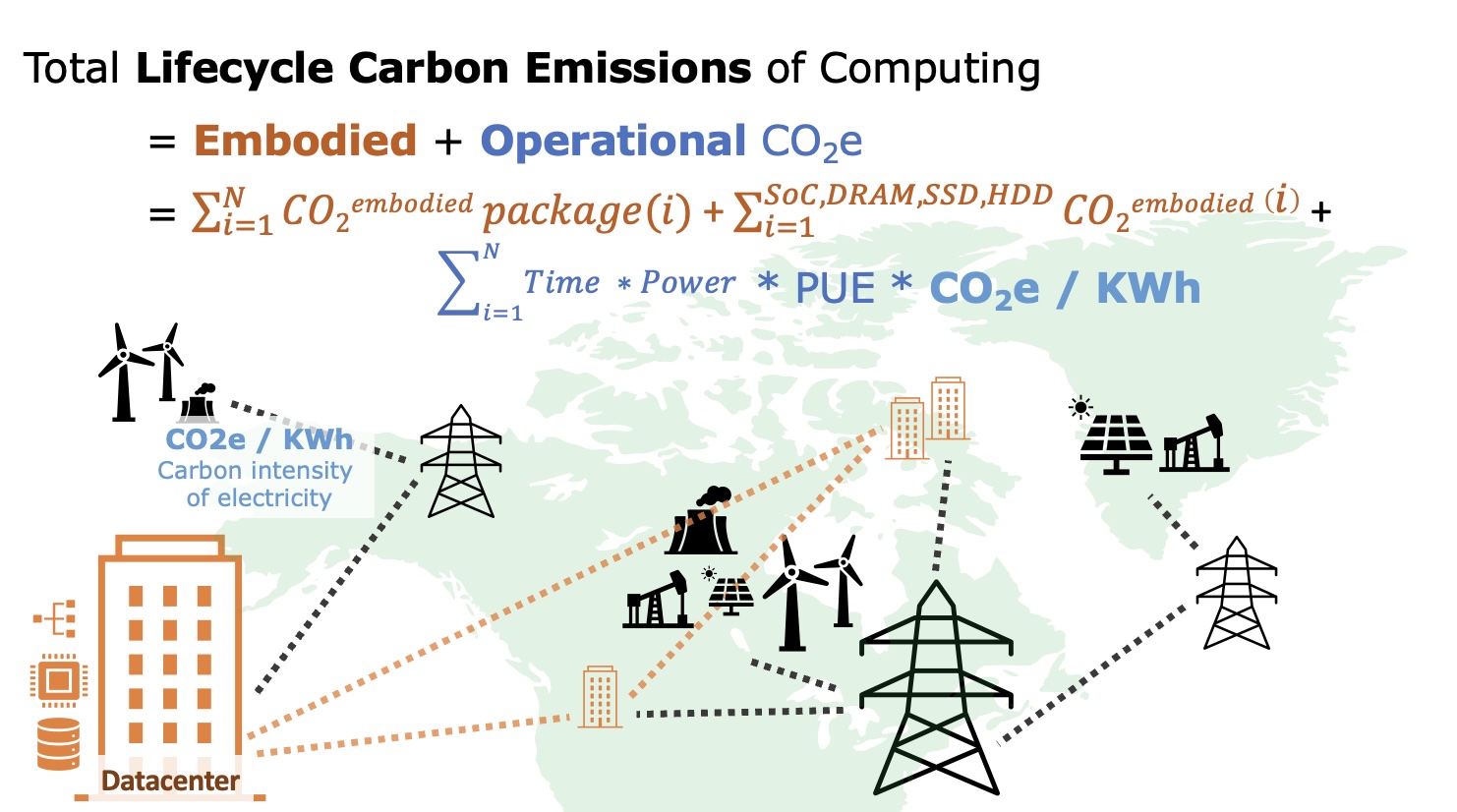}
  \caption{Reducing computing's overall lifecycle carbon emissions --- embodied and operational carbon --- needs improvements from multiple sources: (1) computers designed with minimal embodied carbon and manufactured with lower chemical gasses, and (2) energy efficient hardware that runs on datacenters with optimized PUE and sourced with electricity of lower carbon intensity.}
  \label{fig:lifecycle-emission-reduction}
\end{figure}

\noindent \textbf{Sustainability as a Computer System Design Principle:}
Figure~\ref{fig:lifecycle-emission-reduction} depicts sources for carbon footprint improvement of warehouse-scale computing.  
With carbon telemetry, high-quality carbon datasets and metrics, it opens new design and optimization opportunities across the system stack, from workload management in warehouse-scale computing infrastructures, programming languages and runtime management to system architecture and hardware design. 
At the datacenter level, what new features need to be introduced to WSC infrastructures, for datacenter operators to cooperate with power grids and reduce its operational carbon footprint? Recent endeavors are already exploring carbon-aware datacenter design and management~\cite{hotcarbon-acun}, with advanced planning~\cite{xing2023carbon,Lin_2023} or in real-time~\cite{radovanovic2021carbon}. 

In addition, we envision significant potential for operational carbon footprint reduction in datacenter computing if delay tolerance of computations is visible. At a coarser-granularity, services can be designed with feature tiers in mind. By embedding modularity into cloud services, the power consumption of the services at the datacenter scale can be modulated seamlessly depending on power capacity availability or carbon intensity of electricity. At a finer-granularity, application software can be designed with energy and carbon in mind to enable design space tradeoff between execution time performance and carbon emissions, such as Treehouse~\cite{anderson2022treehouse}.

The design space that puts carbon as the first-class optimization principle brings new challenges. 
\begin{itemize}
    \item Higher energy efficiency does not always translate into lower carbon emissions.  Carbon intensity of electricity varies over time and depends on \textit{how} electricity is generated at power grids in different geographic regions.
    \item The supply chain of electronics manufacturing spans across many geographic regions of distinct energy generation characteristics and over time, leading to a wide design space for \textit{greener} semiconductor manufacturing.
    \item  
Optimizing the efficiency of semiconductor manufacturing process and improving the chemical gas abatement process are effective ways to further reduce the embodied carbon emissions of AI systems.
\end{itemize}

\section{Conclusion}
To ensure an environmentally-sustainable growth of AI, we must focus on efficiency optimization holistically across the entire AI system stack --- data, algorithms and models, systems, and infrastructures at-scale. Efficiency optimization helps bend the ever-increasing resource demands of AI. 
AI also plays an important role in the solution space, demonstrating significant potential to discover new catalysts to address energy storage efficiency challenges~\cite{zitnick2020introduction},  to unlock the potential of renewable energy generation~\cite{acun2023unlocking}, to accelerate the discovery of greener chemical gasses for semiconductor manufacturing abatement, and many more. Furthermore, a circular economy for computing~\cite{ocp-2021}, from consumer electronics to infrastructures at-scale, that supports the sustainability principle of reduce, reuse, repair and recycle, will have a step-function change to reducing computing's environmental impact. 
Looking forward, sustainability for computing means more than delivering first-class computing capabilities with minimal impact on the environment --- large-scale computing infrastructures need to become more reliable, secure and resilient to extreme weather events~\cite{wu2021sociotechnological}.
It is upon us --- each and everyone of us --- to contribute to a sustainable future for computing and the society. 

\section*{Acknowledgements}
This work is based on the experience and key lessons learned during the Green AI journey that many colleagues at Meta have contributed to. The original paper on Sustainable AI: Environmental Implications, Challenges and Opportunities is published at the 2022 Conference on Machine Learning and Systems~\cite{wu2022sustainable}. 
In addition, we would like to highlight the important efficiency research being carried out over the overall AI model development cycle by Newsha Ardalani, Zachary DeVito, Mostafa Elhoushi, Jeff Johnson, Samual Hsia, Basil Hosmer, Michael Kuchnik, and Yejin Lee. We would also like to thank our partners: Justin Meza, Raghu Prabhu, Thote Gowda, Katherine Hurrell, Ricky Ghoshroy, Bruce McLeish, Brian White, Janaki Vamaraju, Alex Bruefach, Tobias Tiecke, Jordan Tse, Frances Amatruda, Sylvia Lee, Nikky Avila, Holly Lahd, Aynsley Kretschmar, Urvi Parekh, Brent Morgan, Peter Freed, Jeff Bladen for their inputs and collaboration on energy analytics and carbon emission modeling; Doug Carmean and Larry Zitnik for brainstorming optimization opportunities at the intersection of AI, computing, and sustainability; Joelle Pineau for her leadership and vision without which this work would not have been possible.



\bibliographystyle{IEEEtranS}
\bibliography{refs}

\end{document}